\documentclass[a4paper, 10pt, conference]{ieeeconf}      
\usepackage{FG2020}

\FGfinalcopy 

\IEEEoverridecommandlockouts                              
\usepackage{graphics} 
\usepackage{epsfig} 
\usepackage{mathptmx} 
\usepackage{times} 
\usepackage{amsmath} 
\usepackage{amssymb}  

\def\FGPaperID{9} 
\usepackage{mathtools}
\usepackage{balance}
\usepackage{epsfig}

\title{\LARGE \bf
Recognizing Families through Images with Pretrained Encoder
}

\author{\parbox{16cm}{\centering
    {\large Tuan-Duy H. Nguyen, Huu-Nghia H. Nguyen, and Hieu Dao}\\
    {\normalsize
    Faculty of Information Technology, University of Science, VNU-HCM, Ho Chi Minh City, Vietnam\\}}
}

\begin{document}

\ifFGfinal
\thispagestyle{empty}
\pagestyle{empty}
\else
\author{Anonymous FG2020 submission\\ Paper ID \FGPaperID \\}
\pagestyle{plain}
\fi
\maketitle

\begin{abstract}

Kinship verification and kinship retrieval are emerging tasks in computer vision. Kinship verification aims at determining whether two facial images are from related people or not, while kinship retrieval is the task of retrieving possible related facial images to a person from a gallery of images. They introduce unique challenges because of the hidden relations and features that carry inherent characteristics between the facial images. We employ 3 methods, FaceNet, Siamese VGG-Face, and a combination of FaceNet and VGG-Face models as feature extractors, to achieve the 9th standing for kinship verification and the 5th standing for kinship retrieval in the Recognizing Family in The Wild 2020 competition. We then further experimented using StyleGAN2 as another encoder, with no improvement in the result.

\end{abstract}

\section{Introduction}
Over the last few years, the application of computer vision in kinship verification has been gaining attention with many benchmark datasets released by several research groups such as KinshipW \cite{kinshipw}, UB Kinface 2.0 \cite{ubkinface2}, Family101 \cite{famili101}, Families In the Wild \cite{Robinson2018b}. The goal of kinship verification is to determine if two people are related to each other through some kin relationship (e.g., father-son), using only facial features retrieved from images. Inputting a pair of facial images and the output is a binary value of 0 or 1 correspond to the relationship being \emph{non-kin} or \emph{kin}.

Kinship search and retrieval is essentially an extension of kinship verification in the sense that it is a many-to-many problem rather than one-to-one. The goal of this task is to retrieve a ranked list of family member given a set of photos from a subject person.  More formally, given two sets of facial images probe and gallery of sizes $K$ and $N$, respectively. Each subject in the probe set contains one or more images of the subject person. The output is a $K \times N$ matrix, in which the $j$-th cell on the $i$-th row denotes the gallery index of the rank $j$-th photo in the query of the $i$-th subject.

This particular application of computer vision is worth investigating because it can play a crucial role in forensic science and other pressing tasks \cite{Qin2019}. One notable example is the application of looking for missing children after many years. In such cases, it is impossible to have a direct reference to people in search. With visual kinship techniques, we can instead rely on the similarity between the features of parents to that of candidates, in their current appearance, to look for them with surveillance databases.

In this research, our main concerns are with two problem statements: kinship verification and kinship search and retrieval, which correspond to Track I and Track III of the Recognizing Family in The Wild (RFIW) 2020 competition, respectively. 


The rest of the paper is organized as follows: Section \ref{literature-review} review previous works in the realm of visual kinship, in particular methods involving metric learning; Section \ref{methods} describes methods correspond to our submissions in the RFIW 2020 competition; Section \ref{evaluation} contains information about the RFIW 2020 dataset as well as the result of discussed methods; Section \ref{conclusion} concludes our work and suggests possible improvements.

\section{Literature review} \label{literature-review}
Early attempts in kinship verification task focus on learning the manual engineered features from the images and then make inferences based on these features, with learning schemes such as spatial pyramid learning-based  \cite{zhou2011}. The authors in \cite{somanath} propose an ensemble learning model for kinship tasks with several kernels to find clusters among the examples and to ensure smaller dissimilarity between kin samples compared to non-kin samples.  The authors in \cite{Lu-2014b} introduced a new Neighborhood Repulsed Metric Learning (NRML) method for kinship verification, aiming at learning a novel metric distance that maximizes interclass (non-kin samples) distance while minimizing intraclass distance. 

Another trend of approaches is based on treating kinship verification as a multi-view problem based on the multiple features that complement each other. The authors in \cite{Lu-2014b,hu2014} introduced learning multiple similarity metrics. Similarly,  the authors in \cite{zhao2018learning} proposed Multiple Kernel Similarity Metric (MKSM) for kinship verification by using a weighted combination of similarities to introduce feature fusion. However, these approaches suffer from losing some properties from each view when project multiple presentations of the data into a metric space. To mitigate this problem, the authors in \cite{hu2017a} suggested a new method that merges common information among the features while preserving the information in each view. Recently, the authors in \cite{transfer-learning-feature-fusion} proposed a framework to efficiently select the deep visual features in the kinship verification task using a three-stage procedure. The authors in \cite{SILD+WCCN} introduced a new framework named Side-Information based Linear Discriminant Analysis integrating Within Class Covariance Normalization, which involves projecting deep visual features through a special discriminative subspace proposed by the method.

What most of the above approaches are assuming is that kinship can be learned by using symmetric distance. In fact, there are also asymmetric relations in kinship (e.g., father vs. daughter in terms of age and gender). Therefore,  the authors in \cite{madpod2017} propose utilizing both the asymmetric distance and the symmetric distance for kinship verification. 

With the explosion in popularity of neural networks, many recent approaches also employ deep structures. For instance, CNNs are used as descriptors \cite{zhang2015a} or used in conjunction with metric learning by distilling the information before applying metrics \cite{lu2017discriminative}. Another direction is based on hierarchical facial representation, where the authors in \cite{kohli2016} use anthropological facial features for Deep Belief Networks feature extraction. Generative adversarial networks also help materializing intuition, such as comparing potential children images generated from parents with actual children  \cite{ozkan2018kinshipgan}.

However, it is worth noticing that deep techniques might come with severe defects. Notably, the authors in \cite{Dawson2019} and \cite{comments-kfiw} point out that current visual kinship benchmarks can be exploited with the "from same photograph" (FSP) features. This is due to the fact that researchers construct these benchmarks with cropped images of family photos. Hence, the impressive performance of deep networks might have a significant attribution for their ability to extract FSP features.




\section{Methods}
\label{methods}

\subsection{Inception-ResNet-v1 with Triplet Loss from FaceNet}

\begin{figure}[htbp]
    \centering
    \includegraphics[width=0.4\textwidth,keepaspectratio]{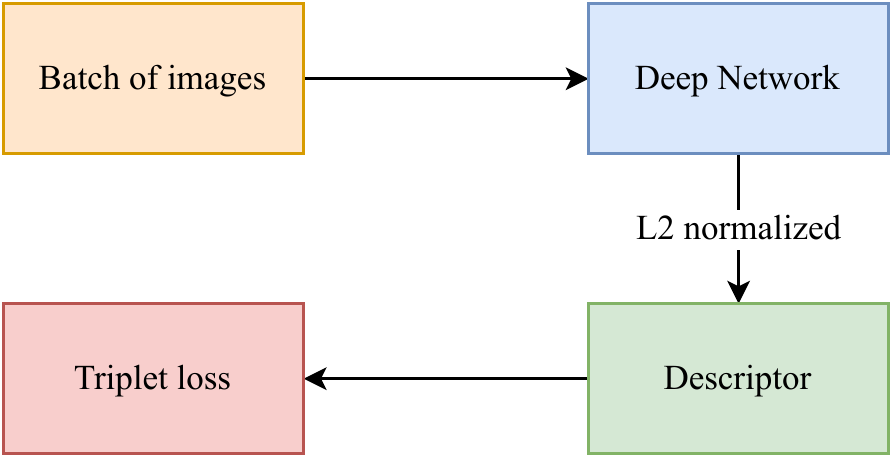}
    \caption{FaceNet model structure.  \cite{Schroff2015}}
    \label{fig:facenet-structure}
\end{figure}

FaceNet \cite{Schroff2015} is a representation learning strategy for face images. It aims to find an embedding $f(x)$ so that the squared distance between images from a similar group is small, while the squared distance between different groups is large. In the context of kinship recognition, the idea is to minimize the distance between people coming from the same family while increasing the distance between people with no kin relation. The implementation is illustrated in Fig. \ref{fig:facenet-structure}. 

FaceNet employs Triplet Loss as the main loss function in the training phase. 

$$L = \sum_{i}^{N}\left[\left\|f\left(x_{i}^{a}\right)-f\left(x_{i}^{p}\right)\right\|_{2}^{2}-\left\|f\left(x_{i}^{a}\right)-f\left(x_{i}^{n}\right)\right\|_{2}^{2}+\alpha\right]_{+}$$

with $x_i^a$ is the anchor image, $x_i^p$ is the positive image, and $x_i^n$ is the negative image.

The main incentive behind this choice is that this loss allows images from the same group to exist in a manifold while still ensuring that images from other groups are away by a margin $\alpha$.

In this run of experiments, we facilitate Inception-ResNet-v1 \cite{Langkvist2014} as the core deep architecture of FaceNet for the sake of quick implementation. 



A facial image going through our system is first L2-normalized before its feature vector is extracted by the FaceNet feature extractor. After that, the FaceNet loss between the two images is calculated, giving us the distance between a pair of images in the test dataset. In order to convert the distance into a probability score, we use cumulative distance (CD) and inverse distance weighting (IDW) \cite{shepard1968-} as interpolation functions. However, using interpolation to convert the distance into a probability score is not a robust method if the data is not evenly distributed.

\subsection{ResNet-50v2 with Additive Angular Margin Loss from ArcFace}

Similar to FaceNet, ArcFace  \cite{Deng2018} is not an architecture or a family of architectures, but a family of loss functions that aim to discriminate the latent representation of deep neural networks. Although softmax-based and triplet loss-based methods obtain excellent results on many benchmarks \cite{Deng2018, Schroff2015}, there are studied drawbacks that need to overcome. For the softmax loss: (1) the size of the linear transformation matrix $W \in \mathbb{R}^{d \times n}$ increase linearly with growing $n$; (2) the learned features are only separable for the closed dataset and not discriminative enough for open face recognition problems. For the triplet loss: (1) there is an explosion in the number of face triplets on large datasets, which directly correspond to the significant growth in the number of iteration steps; (2) sample mining is quite a difficult problem for effective training.

The primary idea of ArcFace is to correspond the dissimilarity of feature vectors to geodesic distance on a hypersphere. Particularly, the training objective becomes maximizing the geodesic margin between the sample and representative centers. Based off of softmax and the angular margin discussed in SphereFace  \cite{Liu2017} and CosFace  \cite{Wang2018}, the loss function of ArcFace is as follows:

$$L = -\frac{1}{N} \sum^N_{i=1}log\frac{e^{s(cos(\theta_{y_i}+m))}}{e^{s(cos(\theta_{y_i}+m))} + \sum^n_{j=1, j \neq y_i}e^{s cos\theta_j}}$$

where there are $N$ samples in a batch, each sample $x_i$ belongs to class $y_i$, $\theta_{y_i}$ is the angle between the $j$-th column of the weight $W \in \mathbb{R}^{d \times N}$, $d$ is the size of the embedding feature, usually set to 512, $m$ is the additive angular margin penalty between $x_i$ and $W_{y_i}$. The embedding feature $\|x_i\|$ is also L2-normalized and then rescale to $s$ \cite{Deng2018}.

The ArcFace loss lent its numerical resemblance from SphereFace and CosFace as all three margin penalties correspond to the angular margin on a hypersphere. In fact, ArcFace improves on SphereFace by using the arc-cosine function on the dot product between the deep features and weight normalization for the angular logit $\theta_j$, instead of angular approximation, hence stabilizing the training without the need of a hybrid loss function. It also improves the discriminative power of CosFace by adding an additive angular margin $m$ to the angle before obtaining the target logit by the cosine function and rescaling all logits.

As kinship tasks resemble clustering tasks, we use a pretrained ResNet-50v2  \cite{He2016} with the maximum ArcFace loss of 0.5 on MS1MV2 \cite{Guo2016} as the feature extractor. Then the similarity of the extracted features is compared pairwise for both competing tasks.

\subsection{Siamese VGG-Face-ResNet-50}

\begin{figure}[htbp]
    \centering
    \includegraphics[width=0.4\textwidth, keepaspectratio]{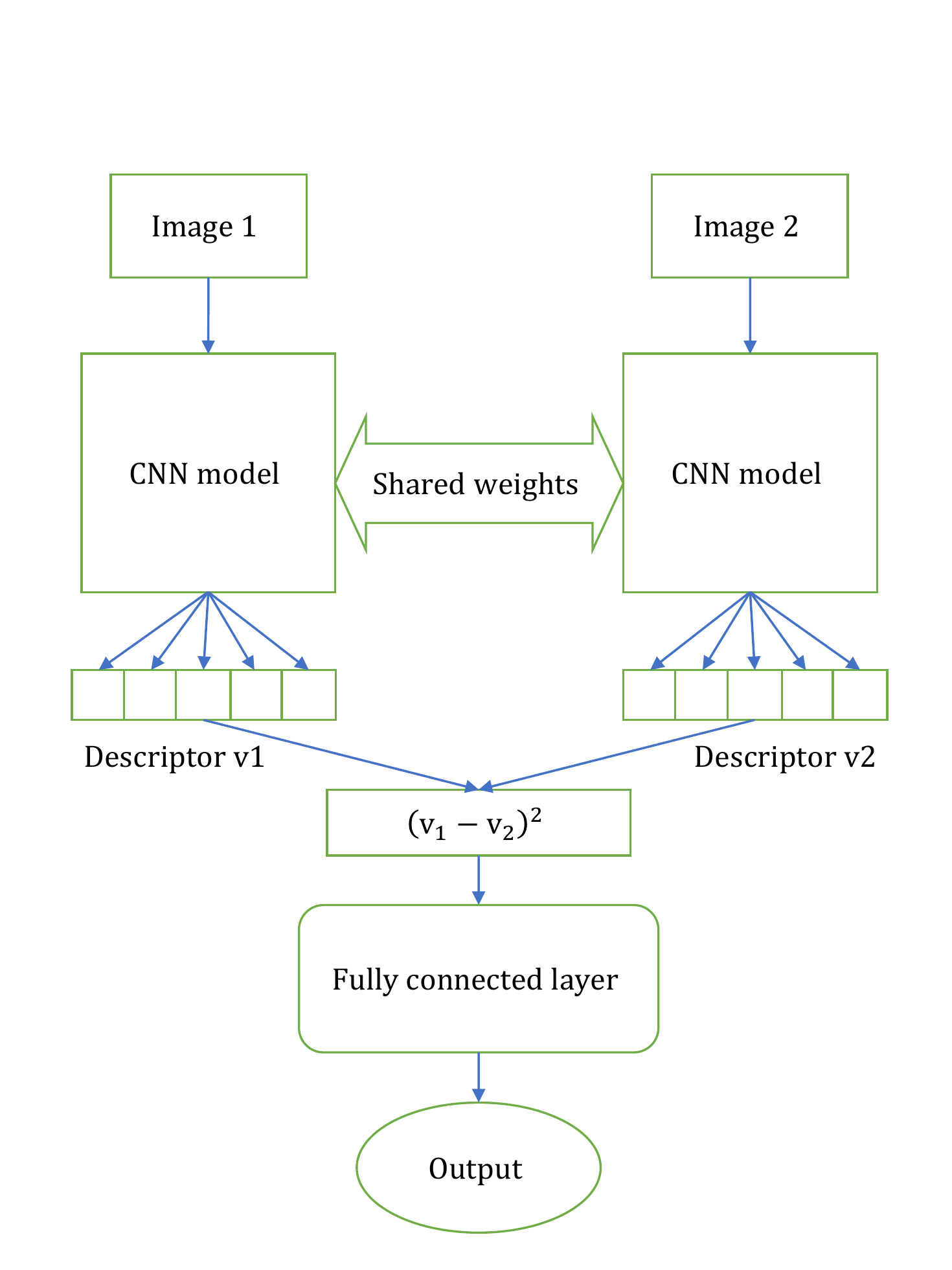}
    \caption{The Siamese network structure.}
    \label{fig:siamese-network-structure}
\end{figure}

Siamese CNN is a deep neural network with two convolutional subnetworks that have the same parameters and structures  \cite{siamese-paper}. The idea behind Siamese CNN is to learn the useful data descriptors that can be used to compare between two input images \cite{siamese-becominghuman}. In kinship context, the Siamese CNN takes a pair of images then predict whether the people in those images are related or not. Adapted from the write-up of Youness Mansar
\footnote{\tiny https://towardsdatascience.com/deep-neural-networks-for-kinship-prediction-using-face-photos-f2ad9ab53834}, in this challenge, we use the VGG-Face-ResNet-50 pretrained model as the convolutional subnetwork model of the Siamese CNN (Fig. \ref{fig:siamese-network-structure}).

The VGG-Face-ResNet-50 is a ResNet-50 model trained on the VGG-Face2 data set  \cite{Parkhi2015}. This is a large-scale face data set with 3.3 million face images and 9000+ identities in a wide range of different ethnicities, accents, professions, and ages. The VGG-Face-ResNet-50 model archive state-of-the-art performance on many famous datasets  \cite{vggface2}. We also use the pretrained model to reduce overfitting and achieve a much faster convergence rate, especially if the source task and target task are close.

For each image as the input, the output of the VGG-Face-ResNet-50 model is a 2048-length vector. After a layer of $L_2$ normalization, this vector becomes a feature vector. For a pair of images, we have two descriptors $v_1$ and $v_2$, using for calculating the distance between two images by the square of the difference between $v_1$ and $v_2$. Finally, the distance is fed into a fully connected layer for classification.

\subsection{FaceNet and VGG-Face joint descriptor}
 This is an engineering exploitation of previously discussed feature descriptors, implemented according to \emph{mattemilio} winning solution\footnote{\tiny https://www.kaggle.com/c/recognizing-faces-in-the-wild/discussion/103670\#latest-596562}. Particularly, extracted features from VGG-Face-ResNet-50 and FaceNet are enhanced by concatenating different vector combination: $x_1 + x_2$, $x_1 - x_2$, $x_1 \times x_2$, $\sqrt{x_1}+\sqrt{x_2}$, and $x_1^2 + x_2^2$, with $x_1$ and $x_2$ being the feature description embeddings of the input images. Extracted features of pairs from both FaceNet and VGG-Face are concatenated before being processed for verification.
 
 \subsection{Naive StyleGAN2 Encoder}
 There is an interesting constraint that should be imposed on the problem is that family members should be recognized on the basis of physical facial features. Hence, the employed latent features should not escape the space of facial features (e.g., to other feature spaces such as lighting conditions). However, mentioned attempts ignore this constrain and do not employ any facial landmark prior. Hence, our next run of experiments should incorporate a facial landmark map for more concrete performances.
 
 At the same time, StyleGAN2 achieves a remarkable performance in high-quality human face synthesis \cite{karras2019analyzing}. As it is infeasible for us to train a large network as StyleGAN2 from scratch, we employ a pretrained model instead. Particularly, we use the pretrained model of StyleGAN2 on FFHQ, which facilitates a 68-landmark weight mask and VGG16 as the feature extractor \cite{karras2019style}. We hypothesize that the deep capacity of the model, along with the weight mask, should provide a good latent representation of faces.
 
 Since StyleGAN2 is pretrained on FFHQ with high-resolution 1024x1024 images, we scale up the original face images using ImageMagick \footnote{\tiny https://imagemagick.org/} before feeding to the encoder. Due to time and resource constraints, we only encode images of the test set and compute the cosine similarities between test pairs. We then use IDW \cite{shepard1968-} to interpolate the calculated scores of the test set before binarizing them by rounding.

\section{Evaluation}
\label{evaluation}
\subsection{Dataset}



Our datasets for two tasks are subsets of the FIW Dataset  \cite{Robinson2018b} with over 13,000 family photos of 1,000 family trees with 4-to-38 members. Interestingly, the dataset is released under several versions for different competitions. This work is produced in response to the Recognizing Families in the Wild 2020 (RFIW2020) competition and is evaluated using this specific distribution of the FIW dataset. We participated with handles \textbf{danbo3004} and \textbf{huunghia160799}.

RFIW2020 is a competition held in conjunction with the 14th IEEE International Conference on Automatic Face and Gesture Recognition (FG2020)\footnote{\tiny https://web.northeastern.edu/smilelab/rfiw2020/}. The competition supports 3 competing tracks with different data splits for each Track:
\begin{enumerate}
    \item Kinship verification (one-to-one)
    \item Tri-subject verification (one-to-two)
    \item Search and Retrieval (many-to-many)
\end{enumerate}

Our experiments concern only Track I and Track III of the competition with the following distribution:
\begin{itemize}
    \item Track I: Kinship verification:
        \begin{itemize}
            \item \emph{Train:} 21920 images from 571 families; 6983 pairs of relationship
            \item \emph{Validation:} 5045 images from 192 families; 71584 pairs of relationship
            \item \emph{Test:} 5226 images; 39742 queries
        \end{itemize}
    \item Track III: Search and retrieval:
        \begin{itemize}
            \item \emph{Train:} 20924 images from 562 families
            \item \emph{Validation:} 334 images from 61 families for probes; 5116 images from 192 families for the gallery
            \item \emph{Test:} 1487 probe images; 3897 gallery images
        \end{itemize}
\end{itemize}

\subsection{Environment setup}

For preparing the submission to the test server of RFIW2020, we employ Google Colaboratory with the following specifications:
\begin{itemize}
    \item CPU: 1 x single core hyperthreaded (i.e., 1  core, 2 threads) Xeon Processors @2.3Ghz (No Turbo Boost) , 45MB Cache
    \item GPU: 1 x NVDIA Tesla T4
    \item RAM: 25GB
\end{itemize}
\subsection{Experimental results}

\subsubsection{Kinship verification}
    \begin{table}
       \caption{Track I results}
        \centering
        \begin{tabular}{|l|l|}
            \hline
            Experiment & Avergage Acc. \\ \hline
            StyleGAN2 Encoder &  0.548\\ \hline
            ResNet-50v2-ArcFace & 0.563 \\ \hline
            Inception-ResNet-v1-FaceNet-CD & 0.681   \\ \hline
            Inception-ResNet-v1-FaceNet-IDW & 0.697 \\ \hline
            Siamese VGG-Face & 0.703 \\ \hline
            Siamese Inception-ResNet-v1-FaceNet & 0.720 \\ \hline
            \textbf{FaceNet+VGG-Face} &  \textbf{0.730}      \\ \hline
        \end{tabular}%
    \label{tab:track1}
    \end{table}
    
    \begin{figure}
        \centering
        \includegraphics[width=\linewidth]{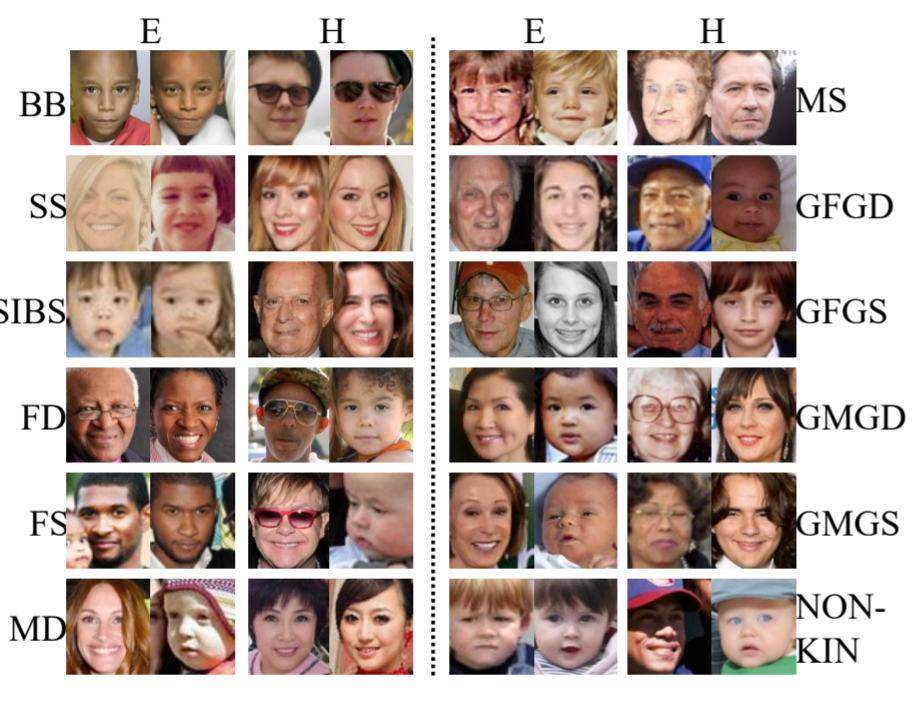}
        \caption{Sample pairs for the categories of Track 1. For each, sample pairs with similarity scores near the threshold (i.e., hard (H) samples), along with highly confident predictions (i.e., easy (E) samples). \cite{robinson2020recognizing}}
        \label{fig:track1-samples}
    \end{figure}

    For the task of kinship verification, we simply perform binary classification between pairs of faces \ref{fig:track1-samples}. For the ArcFace experiment, we compute the cosine similarity between extracted features and impose a hard threshold of $0.6$ to classify kin and non-kin pairs. In the FaceNet experiment, we use a threshold of $0.5$ for the probability produced by cumulative distance and inverse distance weighting and get similar performance. The Siamese VGG-Face and VGG-Face + FaceNet experiments use an end-to-end model with softmax loss as the last layer for classification. Within the competition time frame, our best result is achieved using VGG-Face + FaceNet and stand at the 9th rank on the public leaderboard. Our naive attempt to use StyleGAN2 after the competition does not yield an expected improvement. We postulate this is due to the lack of a proper classification. Details are reported in Tab. \ref{tab:track1}.

\subsubsection{Search and Retrieval}
    \begin{table}
    \caption{Track III results}
     \centering
        \begin{tabular}{|l|l|}
            \hline
            Experiment & MAP \\ \hline
            ResNet-50v2-ArcFace & 0.280 \\ \hline
            Siamese VGG-Face & 0.290\\ \hline
            \textbf{Inception-ResNet-v1-FaceNet} & \textbf{0.316}\\ \hline
        \end{tabular}%
        \label{tab:track3}
    \end{table}
    
    \begin{figure}
        \centering
        \includegraphics[width=\linewidth, keepaspectratio]{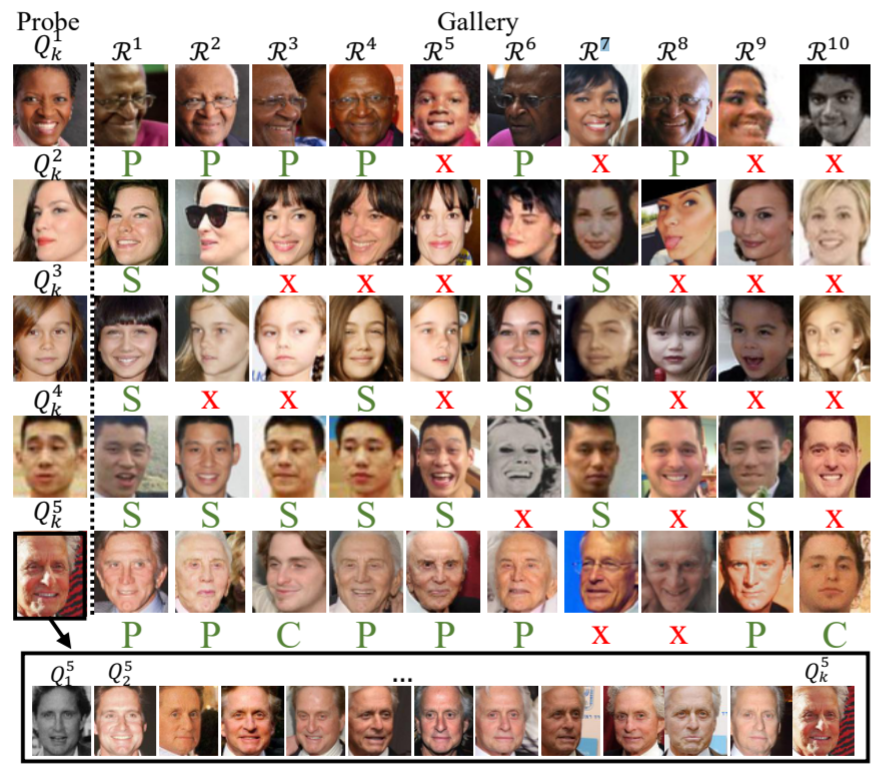}
        \caption{Track 3 sample results (Rank 10). For each query (row) one or more faces of the probe returned the corresponding samples of gallery as top 10. x (red) depicts false predictions. True predictions display the relationship type ( green): P for parent; C for child; S for sibling. \cite{robinson2020recognizing}}
        \label{fig:track3-samples}
    \end{figure}
    
    
    For the task of search and retrieval, we simply compute the pairwise cosine similarity between each probe and the whole gallery and sort the gallery by similarity with respect to the probe. We adapt the ArcFace and VGG-Face experiments from Track I with similar settings. Within the competition time frame, our best result in the competition is a MAP of 0.290, achieved using Siamese VGG-Face and stand at the 5th rank on the public leaderboard. We later adapt the Inception-ResNet-v1-FaceNet after the competition and achieves an improved MAP result of 0.316. Details performance are reported in Tab. \ref{tab:track3}.

\balance
\section{Conclusion and future work}
\label{conclusion}

Kinship-related tasks in computer vision, such as kinship verification and kinship retrieval, are gaining popularity as they propose unique and interesting domain problems. In this work, we evaluate a variety of encoders based methods using existing models as feature extractors on the RFIW2020 dataset. Among these methods, FaceNet \cite{Schroff2015} combining with VGG-Face \cite{vggface2} descriptors show the best result for kinship verification. Along with that, Inception-ResNet-v1-FaceNet yields the best result for kinship search and retrieval. To prevent the overkill capacity of deep neural networks, which might undesirably employ FSP features, we attempted using StyleGAN2, which incorporates a facial landmark mask for anthropological features. However, this does not yield an expected improvement and needs further investigation.

\section{Acknowledgement}
We would like to thank Assoc. Prof. Minh-Triet Tran for reviewing and giving feedbacks to this work.

\bibliographystyle{ieee}
\bibliography{fiw, additional}

\begin{thebibliography}{10}\itemsep=-1pt

\bibitem{comments-kfiw}
M.~Bordallo~Lopez, E.~Boutellaa, and A.~Hadid.
\newblock Comments on the " kinship face in the wild " data sets.
\newblock {\em IEEE Transactions on Pattern Analysis and Machine Intelligence},
  38, 02 2016.

\bibitem{Dawson2019}
M.~Dawson, A.~Zisserman, and C.~Nell{\aa}ker.
\newblock {From Same Photo: Cheating on Visual Kinship Challenges}.
\newblock pages 654--668. sep 2019.

\bibitem{Deng2018}
J.~Deng, J.~Guo, N.~Xue, and S.~Zafeiriou.
\newblock {ArcFace: Additive Angular Margin Loss for Deep Face Recognition}.
\newblock (1), 2018.

\bibitem{transfer-learning-feature-fusion}
F.~Dornaika, I.~Arganda-Carreras, and O.~Serradilla.
\newblock Transfer learning and feature fusion for kinship verification.
\newblock {\em Neural Computing and Applications}, 04 2019.

\bibitem{famili101}
R.~{Fang}, A.~C. {Gallagher}, T.~{Chen}, and A.~{Loui}.
\newblock Kinship classification by modeling facial feature heredity.
\newblock In {\em 2013 IEEE International Conference on Image Processing},
  pages 2983--2987, Sep. 2013.

\bibitem{Guo2016}
Y.~Guo, L.~Zhang, Y.~Hu, X.~He, and J.~Gao.
\newblock {MS-celeb-1M: A dataset and benchmark for large-scale face
  recognition}.
\newblock In {\em Lecture Notes in Computer Science (including subseries
  Lecture Notes in Artificial Intelligence and Lecture Notes in
  Bioinformatics)}, volume 9907 LNCS, pages 87--102, 2016.

\bibitem{He2016}
K.~He, X.~Zhang, S.~Ren, and J.~Sun.
\newblock {Deep residual learning for image recognition}.
\newblock In {\em Proceedings of the IEEE Computer Society Conference on
  Computer Vision and Pattern Recognition}, volume 2016-Decem, pages 770--778,
  2016.

\bibitem{siamese-becominghuman}
B.~Holländer.
\newblock Siamese networks: Algorithm, applications and pytorch implementation.

\bibitem{hu2017a}
J.~{Hu}, J.~{Lu}, and Y.~{Tan}.
\newblock Sharable and individual multi-view metric learning.
\newblock {\em IEEE Transactions on Pattern Analysis and Machine Intelligence},
  40(9):2281--2288, Sep. 2018.

\bibitem{hu2014}
J.~Hu, J.~Lu, J.~Yuan, and Y.-P. Tan.
\newblock Large margin multi-metric learning for face and kinship verification
  in the wild.
\newblock In D.~Cremers, I.~Reid, H.~Saito, and M.-H. Yang, editors, {\em
  Computer Vision -- ACCV 2014}, pages 252--267, Cham, 2015. Springer
  International Publishing.

\bibitem{siamese-paper}
Y.~L. E.~S. Jane~Bromley, Isabelle~Guyon and R.~Shah.
\newblock Signature verification using a "siamese" time delay neural network.
\newblock 1994.

\bibitem{karras2019style}
T.~Karras, S.~Laine, and T.~Aila.
\newblock A style-based generator architecture for generative adversarial
  networks.
\newblock In {\em Proceedings of the IEEE Conference on Computer Vision and
  Pattern Recognition}, pages 4401--4410, 2019.

\bibitem{karras2019analyzing}
T.~Karras, S.~Laine, M.~Aittala, J.~Hellsten, J.~Lehtinen, and T.~Aila.
\newblock Analyzing and improving the image quality of stylegan.
\newblock {\em arXiv preprint arXiv:1912.04958}, 2019.

\bibitem{kohli2016}
N.~Kohli, M.~Vatsa, R.~Singh, A.~Noore, and A.~Majumdar.
\newblock Hierarchical representation learning for kinship verification.
\newblock {\em IEEE Transactions on Image Processing}, 26:1--1, 09 2016.

\bibitem{SILD+WCCN}
O.~Laiadi, A.~Ouamane, A.~Benakcha, A.~Taleb-Ahmed, and A.~Hadid.
\newblock Learning multi-view deep and shallow features through new
  discriminative subspace for bi-subject and tri-subject kinship verification.
\newblock {\em Applied Intelligence}, pages 1--15, 2019.

\bibitem{Langkvist2014}
M.~L{\"{a}}ngkvist, L.~Karlsson, and A.~Loutfi.
\newblock {Inception-v4, Inception-ResNet and the Impact of Residual
  Connections on Learning}.
\newblock {\em Pattern Recognition Letters}, 42(1):11--24, 2014.

\bibitem{Liu2017}
W.~Liu, Y.~Wen, Z.~Yu, M.~Li, B.~Raj, and L.~Song.
\newblock {SphereFace: Deep hypersphere embedding for face recognition}.
\newblock {\em Proceedings - 30th IEEE Conference on Computer Vision and
  Pattern Recognition, CVPR 2017}, 2017-January:6738--6746, 2017.

\bibitem{kinshipw}
J.~{Lu}, J.~{Hu}, V.~E. {Liong}, X.~{Zhou}, A.~{Bottino}, I.~U. {Islam}, T.~F.
  {Vieira}, X.~{Qin}, X.~{Tan}, S.~{Chen}, S.~{Mahpod}, Y.~{Keller},
  L.~{Zheng}, K.~{Idrissi}, C.~{Garcia}, S.~{Duffner}, A.~{Baskurt},
  M.~{Castrillón-Santana}, and J.~{Lorenzo-Navarro}.
\newblock The fg 2015 kinship verification in the wild evaluation.
\newblock In {\em 2015 11th IEEE International Conference and Workshops on
  Automatic Face and Gesture Recognition (FG)}, volume~1, pages 1--7, May 2015.

\bibitem{lu2017discriminative}
J.~Lu, J.~Hu, and Y.-P. Tan.
\newblock Discriminative deep metric learning for face and kinship
  verification.
\newblock {\em IEEE Transactions on Image Processing}, 26(9):4269--4282, 2017.

\bibitem{Lu-2014b}
J.~{Lu}, X.~{Zhou}, Y.~{Tan}, Y.~{Shang}, and J.~{Zhou}.
\newblock Neighborhood repulsed metric learning for kinship verification.
\newblock {\em IEEE Transactions on Pattern Analysis and Machine Intelligence},
  36(2):331--345, Feb 2014.

\bibitem{madpod2017}
S.~Mahpod and Y.~Keller.
\newblock Kinship verification using multiview hybrid distance learning.
\newblock {\em Computer Vision and Image Understanding}, 167, 12 2017.

\bibitem{ozkan2018kinshipgan}
S.~Ozkan and A.~Ozkan.
\newblock Kinshipgan: Synthesizing of kinship faces from family photos by
  regularizing a deep face network.
\newblock In {\em 2018 25th IEEE International Conference on Image Processing
  (ICIP)}, pages 2142--2146. IEEE, 2018.

\bibitem{Parkhi2015}
O.~M. Parkhi, A.~Vedaldi, and A.~Zisserman.
\newblock {Deep Face Recognition}.
\newblock (Section 3):41.1--41.12, 2015.

\bibitem{Qin2019}
X.~Qin, D.~Liu, and D.~Wang.
\newblock {A literature survey on kinship verification through facial images}.
\newblock {\em Neurocomputing}, 2019.

\bibitem{vggface2}
W.~X. O. M.~P. Qiong~Cao, Li~Shen and A.~Zisserman.
\newblock {VGGFace2:} a dataset for recognising faces across pose and age.
\newblock 2018.

\bibitem{Robinson2018b}
J.~P. Robinson, M.~Shao, Y.~Wu, H.~Liu, T.~Gillis, and Y.~Fu.
\newblock {Visual Kinship Recognition of Families in the Wild}.
\newblock {\em IEEE Transactions on Pattern Analysis and Machine Intelligence},
  40(11):2624--2637, 2018.

\bibitem{robinson2020recognizing}
J.~P. Robinson, Y.~Yin, Z.~Khan, M.~Shao, S.~Xia, M.~Stopa, S.~Timoner, M.~A.
  Turk, R.~Chellappa, and Y.~Fu.
\newblock Recognizing families in the wild (rfiw): The 4th edition.
\newblock {\em arXiv preprint arXiv:2002.06303}, 2020.

\bibitem{ubkinface2}
J.~L. S.~Xia, M.~Shao and Y.~Fu.
\newblock Understanding kin relationships in a photo.
\newblock {\em IEEE Transactions on Multimedia}, 14(4):1046--1056, 2012.

\bibitem{Schroff2015}
F.~Schroff, D.~Kalenichenko, and J.~Philbin.
\newblock {FaceNet: A unified embedding for face recognition and clustering}.
\newblock In {\em Proceedings of the IEEE Computer Society Conference on
  Computer Vision and Pattern Recognition}, volume 07-12-June, pages 815--823,
  2015.

\bibitem{shepard1968-}
D.~Shepard.
\newblock A two-dimensional interpolation function for irregularly-spaced data.
\newblock In {\em Proceedings of the 1968 23rd ACM National Conference}, ACM
  ’68, page 517–524, New York, NY, USA, 1968. Association for Computing
  Machinery.

\bibitem{somanath}
G.~{Somanath} and C.~{Kambhamettu}.
\newblock Can faces verify blood-relations?
\newblock In {\em 2012 IEEE Fifth International Conference on Biometrics:
  Theory, Applications and Systems (BTAS)}, pages 105--112, Sep. 2012.

\bibitem{Wang2018}
H.~Wang, Y.~Wang, Z.~Zhou, X.~Ji, D.~Gong, J.~Zhou, Z.~Li, and W.~Liu.
\newblock {CosFace: Large Margin Cosine Loss for Deep Face Recognition}.
\newblock {\em Proceedings of the IEEE Computer Society Conference on Computer
  Vision and Pattern Recognition}, pages 5265--5274, 2018.

\bibitem{zhang2015a}
K.~Zhang, Y.~Huang, C.~Song, H.~Wu, and L.~Wang.
\newblock Kinship verification with deep convolutional neural networks.
\newblock pages 148.1--148.12, 01 2015.

\bibitem{zhao2018learning}
Y.-G. Zhao, Z.~Song, F.~Zheng, and L.~Shao.
\newblock Learning a multiple kernel similarity metric for kinship
  verification.
\newblock {\em Information Sciences}, 430:247--260, 2018.

\bibitem{zhou2011}
X.~Zhou, J.~Hu, J.~Lu, Y.~Shang, and Y.~Guan.
\newblock Kinship verification from facial images under uncontrolled
  conditions.
\newblock In {\em Proceedings of the 19th ACM International Conference on
  Multimedia}, MM ’11, page 953–956, New York, NY, USA, 2011. Association
  for Computing Machinery.

\end{thebibliography}

\end{document}